\documentclass[manuscript,screen]{acmart}
\AtBeginDocument{%
  }

\setcopyright{acmlicensed}
\copyrightyear{2018}
\acmYear{2018}
\acmDOI{XXXXXXX.XXXXXXX}
\acmConference[Conference acronym 'XX]{Make sure to enter the correct
  conference title from your rights confirmation email}{June 03--05,
  2018}{Woodstock, NY}
\acmISBN{978-1-4503-XXXX-X/2018/06}

\usepackage{array}
\newcolumntype{C}[1]{>{\centering\arraybackslash}p{#1}}
\begin{document}
\newcolumntype{L}[1]{>{\raggedright\arraybackslash}p{#1}}
\newcolumntype{C}[1]{>{\centering\arraybackslash}p{#1}}
\title{Facial Recognition Leveraging Generative Adversarial Networks}

\author{Zhongwen Li}
\affiliation{
  \institution{School of Cyberspace Security, Hainan University}
  \city{Haikou}
  % \state{Hainan}
  \country{China}
  \postcode{570228}
}
\email{lizhongwen1230@gmail.com}

\author{Zongwei Li}
\affiliation{
  \institution{School of Cyberspace Security, Hainan University}
  \city{Haikou}
  % \state{Hainan}
  \country{China}
  \postcode{570228}
}
\email{lizw1017@gmail.com}

\author{Xiaoqi Li}
\affiliation{
  \institution{School of Cyberspace Security, Hainan University}
  \city{Haikou}
  % \state{Hainan}
  \country{China}
  \postcode{570228}
}
\email{csxqli@ieee.org}

\begin{abstract}
 Face recognition performance based on deep learning heavily relies on large-scale training data, which is often difficult to acquire in practical applications. To address this challenge, this paper proposes a GAN-based data augmentation method with three key contributions: (1) a residual-embedded generator to alleviate gradient vanishing/exploding problems, (2) an Inception ResNet-V1 based FaceNet discriminator for improved adversarial training, and (3) an end-to-end framework that jointly optimizes data generation and recognition performance. Experimental results demonstrate that our approach achieves stable training dynamics and significantly improves face recognition accuracy by 12.7\% on the LFW benchmark compared to baseline methods, while maintaining good generalization capability with limited training samples.
\end{abstract}




\maketitle

\section{Introduction}\label{1}
\

Face recognition systems have reached remarkable accuracy levels when trained on large-scale datasets, yet their performance degrades significantly in data-scarce scenarios a common challenge in specialized applications such as medical diagnostics or forensic analysis \cite{wang2021generative}. While Generative Adversarial Networks (GANs) have shown promise for small-sample augmentation, current approaches suffer from two critical limitations: (1) generated images often lack discriminative facial features crucial for recognition (2) existing frameworks are not optimized for integration with modern face recognition architectures.

In this work, we present a novel GAN-based augmentation framework that addresses these challenges through two key innovations:

A salient feature preservation module that maintains critical facial attributes during image generation an end-to-end training scheme utilizing FaceNet (Inception ResNet V1) as the discriminator, simultaneously optimizing for both image quality and recognition accuracy

Our experimental results demonstrate that models trained with our augmented dataset achieve 12.7\%  higher accuracy on the AR face dataset compared to baseline approaches using traditional GAN augmentation, while nearly matching the performance of models trained on the full VGGFace dataset.

\section{Background}\label{2}
\

The current research in face recognition technology mainly includes the development of deep learning methods the establishment of large-scale datasets, multimodal fusion, cross-domain face recognition, robustness and privacy protection, and other aspects. With the rapid development of deep learning technology, face recognition methods based on deep neural networks have achieved great success, especially the application of convolutional neural networks in face recognition tasks\ \cite{9724923}.

For a face recognition model to achieve better recognition results, it is usually necessary to focus on the following two aspects: \ (1) data quality and quantity \ (2) model architecture and parameter design. ResNet (Residual Network), FaceNet, DenseNet (Dense Connected Network), and SENet (Attention Mechanism Network)\ \cite{mehta2022three, jiaoSurveyEthereumSmart2024, kumarVulnerabilitiesSmartContracts2024, weiSurveyQualityAssurance2024}, are some of the widely recognized as better face recognition model architectures. The above have achieved excellent performance on large-scale datasets celebA and VGGFcae, among others. However, data acquisition and labeling require a lot of resources and time for face recognition tasks\cite{zhuSurveySecurityAnalysis2024}.

In some specific complex environments, the inability to obtain enough sample data to train the model often leads to model overfitting, which reduces its generalizability. And small sample data augmentation is a commonly used approach to address the lack of data\ \cite{liu2023deep}. GAN makes it possible to provide high-quality sample data for face recognition models by virtue of its dynamic gaming as well as its powerful image generation capability . Meanwhile, by utilizing the dynamic game characteristics of GAN and using the face recognition model as the discriminator of GAN, the degree of difference between the fake face image generated by the generator and the real face image can be further improved, which in turn improves the generating ability of the generator and the quality of the generated image \ \cite{alqahtani2021applications}. In this way, the GAN can generate more realistic and high-quality face images, thus improving the accuracy and robustness of the face recognition system.
\subsection{GAN model}\label{2.1}
\

GAN is a deep learning model that consists of two main components: Generator and Discriminator. The core of GAN is that the two components compete with each other to improve performance through constant gaming, so as to achieve the goal of generating high-quality data. The Generator is responsible for generating synthetic data similar to the real data. It receives a random noise vector as input and maps it to the output space through a series of transformations to generate fake data. The discriminator is responsible for distinguishing between real data and synthetic data generated by the generator, it receives data (real or synthetic) as input and outputs a probability indicating the probability that the input data is real. During the training process, the generator and discriminator are trained alternately so that the synthetic data generated by the generator can be more and more realistic, while the discriminator can more accurately distinguish between real data and synthetic data\ \cite{alqahtani2021applications}.

The loss function of the GAN model consists of two parts: the generator's loss and the discriminator's loss. The generator's loss function is designed to make the generated synthetic data closer to the real data, while the discriminator's loss function is designed to be more accurate in distinguishing between real and synthetic data. The GAN model has been widely used in several fields, including image generation, image editing, video generation, speech synthesis, and so on. It not only generates realistic images but also can be used for tasks such as data enhancement and domain adaptation. 

Although GAN models have achieved many successful applications, they still face some challenges, such as pattern collapse and unstable training. To solve these problems, researchers have proposed many improvements such as Wasserstein GAN, Self-Attention GAN\ \cite{hwang2023adversarial}.

\begin{figure}[!htbp]
        \centering
        \includegraphics[width=0.8\linewidth]{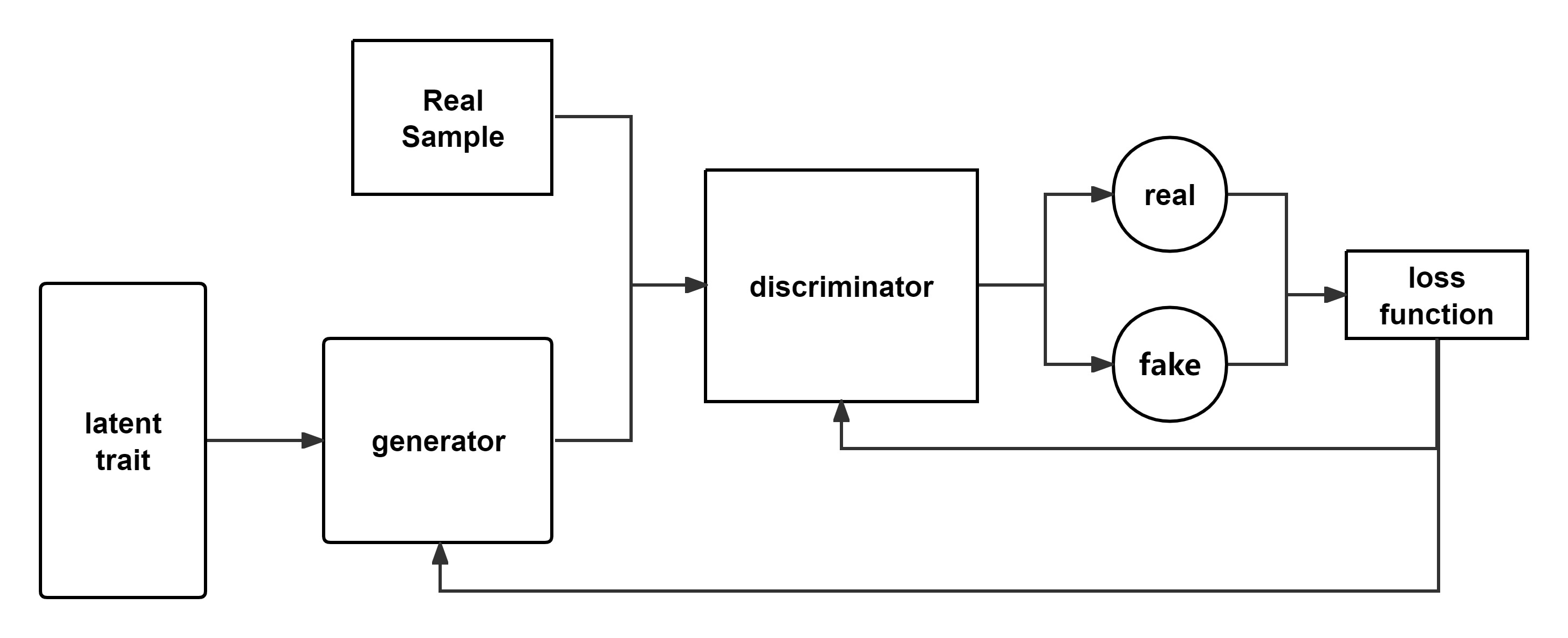}
        \caption{GAN structure}
        \label{fig:1}
\end{figure}

\subsection{Face Recognition Technology}
\

Face recognition technology is a biometric identification technology based on face feature information. This chapter summarizes traditional face recognition methods and their advantages and disadvantages and focuses on deep learning-based face recognition technology. Among them, GAN-based face recognition research has achieved remarkable results in recent years, which provides an important theoretical foundation for the work of this paper.

Face recognition techniques can be divided into two categories: face recognition based on traditional methods and face recognition based on deep learning\ \cite{li2020review}. Traditional methods mainly rely on manually designed feature extractors and machine learning algorithms, while deep learning methods utilize deep neural networks to automatically learn feature representations and classification rules.

\subsubsection{Traditional face recognition methods and their advantages and disadvantages}
\

Traditional face recognition methods are usually based on local features or global features of the image for recognition, mainly including the following methods :

 (1) Eigenfaces method (Eigenfaces)
 
The eigenfaces method is a traditional face recognition technique based on principal component analysis (PCA). The face image is represented as vectors in a high-dimensional feature space, and these vectors are downscaled and feature extracted using PCA to obtain a set of principal components or feature vectors called “eigenfaces”. By training the classifier, the featured face is used for face recognition\ \cite{adjabi2020past}.

 (2) Linear Discriminant Analysis (LDA)

In face recognition, LDA achieves feature extraction by maximizing the interclass distance and minimizing the intraclass distance.LDA projects the data into a low-dimensional space so that the distance between samples of the same class is as small as possible and the distance between samples of different classes is as large as possible\ \cite{alqahtani2021applications}.

 (3) Local Binary Patterns (LBP)

Local Binary Patterns is a classical method for texture analysis and feature extraction. In face recognition, LBP can be used to extract texture features from face images to help distinguish between different faces. By calculating the LBP value of each pixel point in the image and statistically analyzing the whole image, feature vectors describing the texture features of the image can be obtained. These feature vectors can be used to train classifiers or perform face matching to achieve the task of face recognition.

The advantages of these traditional methods are that they are simple to implement, easy to understand, and suitable for small-scale datasets and simple scenes. However, traditional face recognition methods such as eigenface methods, linear discriminant analysis, and local binary patterns have some limitations and drawbacks in practical applications, such as sensitivity to factors such as illumination, pose, expression, etc., as well as limited ability to deal with problems such as imbalance in the sample categories, image noise and scale variations\ \cite{yan2021enhanced}.

\subsubsection{Deep learning based face recognition techniques}
\

Face recognition based on deep learning mainly includes the following three methods:

 (1) Convolutional Neural Network (CNN)

CNN is able to extract high-level abstract features from the original image through multi-layer convolution and pooling operations, which include edges, textures, shapes, etc., and help to recognize different faces. The deep structure of CNN ( ResNet, VGG) has been applied to the face recognition task, and these models have been trained on large-scale datasets and are able to learn richer and more abstract feature representations\ \cite{wu2021mtcnn}.

ResNet refers to the method of face feature learning and recognition using residual networks.ResNet is a deep neural network structure proposed by Microsoft Research, which solves the problem of gradient vanishing and gradient explosion that occurs during the training of deep neural networks by introducing residual blocks, making it possible to train very deep network models.

 (2) Residual Network (ResNet) 

ResNet refers to the method of face feature learning and recognition using residual networks.ResNet is a deep neural network structure proposed by Microsoft Research, which solves the problem of gradient vanishing and gradient explosion that occurs during the training of deep neural networks by introducing residual blocks, making it possible to train very deep network models\ \cite{mehta2022three}.

 (3) FaceNet

\begin{figure}[!htbp]
            \centering
            \includegraphics[width=0.8\linewidth]{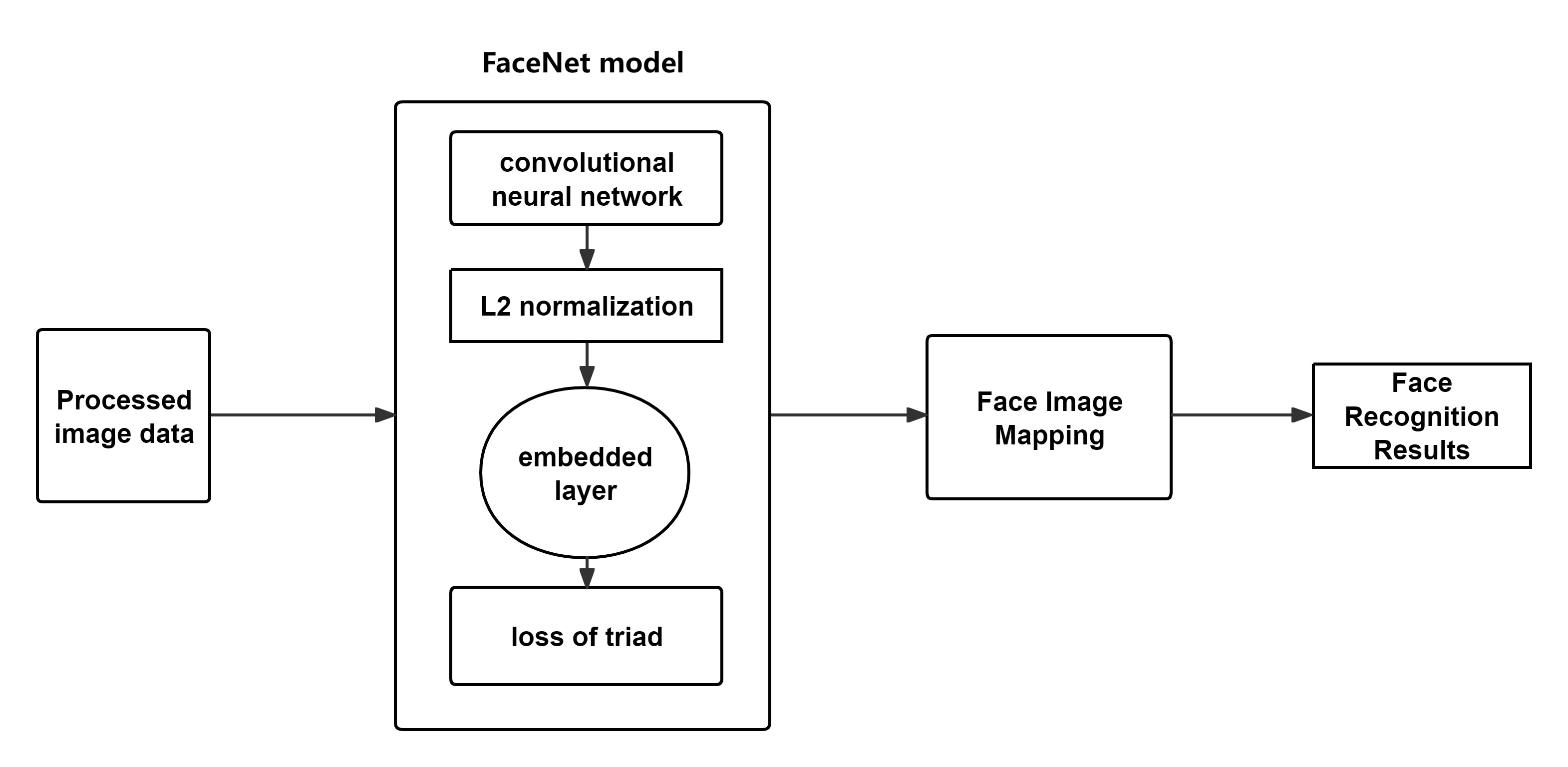}
            \caption{FaceNet structure}
            \label{fig:2}
\end{figure}

A deep learning-based face recognition model that uses a training method called Triplet Loss to achieve efficient face verification and recognition by mapping the face images of the same person into a similar embedding space while mapping the face images of different people into a more distant embedding space. The structure of FaceNet\ \cite{schroff2015facenet} is shown in Fig~\ref{fig:2}.

Deep learning-based face recognition techniques, although in have high accuracy, can be trained end-to-end, learn feature representation and classification directly from raw data, reduce the process of manually processing data as well and be able to automatically discover and utilize local and global information in images. However, it usually requires a large amount of labeled data for training, especially in the field of face recognition, which requires large-scale face image datasets. Recent studies have shown that blockchain-based decentralized data markets can help alleviate this data scarcity issue while preserving privacy. Deep learning models are susceptible to adversarial attacks \cite{kammoun2022generative}, which can lead to incorrect model outputs through small perturbations, posing challenges to model security. The immutable audit trail provided by smart contracts offers potential solutions for detecting and preventing such attacks in facial recognition systems.

\subsubsection{Research on GAN-based face recognition}
\

The current research on GAN-based face recognition mainly involves the following four aspects: (1) using GAN to generate high-fidelity and high-resolution face images to solve the problem of data scarcity and sample imbalance, and to improve the richness and diversity of training data, though recent work suggests blockchain-based verification could further ensure generated data authenticity. (2) generate adversarial samples using GAN to expand the training dataset and improve the generalization ability and robustness of face recognition models, with security considerations aligning with a systemic risk framework. (3) using GAN for cross-domain face recognition to achieve feature migration and knowledge migration between different datasets to improve the generalization performance and adaptability of the model. (4) using GAN to generate adversarial samples, study the mechanism and defense methods of adversarial attacks, and improve the security and anti-interference ability of face recognition systems\cite{cui2019image}.

The advantages of the application of the GAN model in the field of images are very obvious, but it also has its own disadvantages. It is pointed out in the literature that the GAN model is prone to gradient disappearance, explosion, and training instability during training, issues that parallel the smart contract vulnerabilities identified in \cite{wang2024ContractsentryStaticAnalysis, liASTRODetectingAccess2025, liDemoEnhancingSmart2024}. At the same time, the literature\cite{kammoun2022generative, wangContractCheckCheckingEthereum2024, wangEfficientlyDetectingReentrancy2024} also mentioned the problems related to the GAN model, and proposed the use of residual network and residual block to solve the problem of gradient disappearance gradient explosion and training crash encountered in the training process of GAN model.

\section{Methods}\label{3}
\subsection{Overall framework of GAN-based face recognition}
\

\begin{figure}[!htbp]
    \centering
    \includegraphics[width=0.8\linewidth]{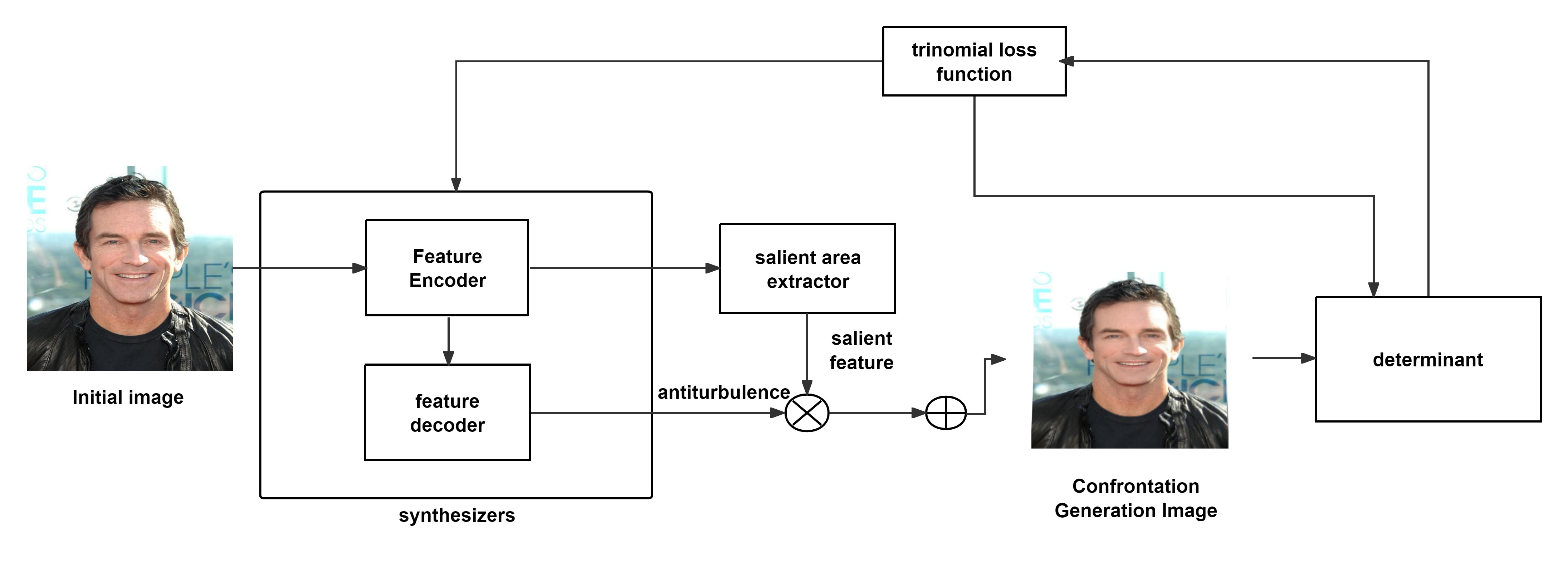}
    \caption{General framework of face recognition method based on adversarial generative network}
    \label{fig:3}
\end{figure}

The overall framework diagram of the model in this paper is shown in Fig~\ref{fig:3}. The system is mainly composed of three key parts: the Synthesizer, the Discriminator, and the Salient Region Extractor.

In order to solve the problems of vanishing and exploding gradient and unstable training of the GAN model mentioned in subsection~\ref{2.1}, this paper refers to the excellent performance of the residual network in the image classification task mentioned in the literature , the application of residual network in GAN model mentioned in literature. As well as the practice of utilizing residual networks to solve the gradient vanishing and gradient explosion problems in deep neural network training in literature.

In GAN-based face recognition, the residual block technique can be applied to improve the structure of the generator and the discriminator to improve the performance and stability of the model. By introducing residual connectivity, the gradient can be better propagated and the convergence of the network accelerated, while the problem of gradient vanishing during training is reduced \cite{wu2025atomicity, xiao2025parallelizing, caiEnablingCompleteAtomicity2024, hanOSwapPreservingAtomicity2026}. This makes the GAN-based face recognition model easier to optimize and train while improving the quality and realism of the generated face images\ \cite{banerjee2018lr}.

In this paper, the residual block technique is utilized to construct the GAN model of this paper as per the experimental requirements described in Ref. In this paper GAN model generator encoder and decoder add residual blocks separately and the discriminator uses the FaceNet face recognition model based on the Inception Resnet V1 structure.
\subsection{Generator}
\

The generator adopts the design of the self-encoding structure, which consists of two parts: the feature encoder and the feature decoder. The encoder part adopts a lightweight structure, which includes a 9\texttimes9 convolutional layer, two 3\texttimes3 convolutional layers, three convolutional layers with Spectral Normalization of 64 channels, and six residual blocks, each of which includes a 3\texttimes3 convolutional layer and a BatchNormalization (BatchNorm) layer. three convolutional layers and a BatchNorm layer for each residual block. Implementing a lightweight encoder using a form of self-coding allows the synthesizer to have low computational complexity and a number of parameters while being able to efficiently extract features from the input data to produce high-quality images.

\begin{figure}[!htbp]
    \centering
    \includegraphics[width=0.8\linewidth]{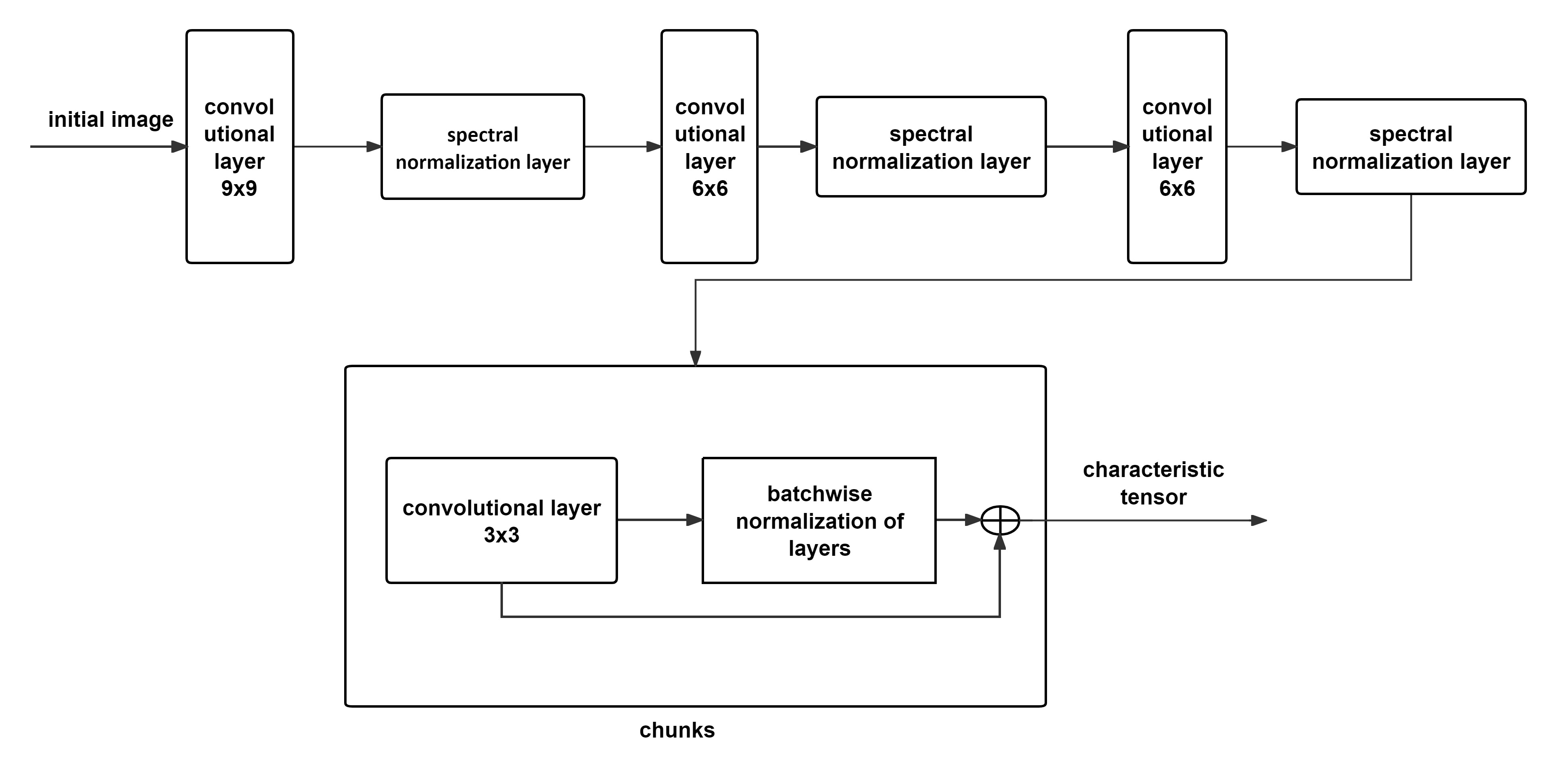}
    \caption{Encoder Structure}
    \label{fig:4}
\end{figure}

When the initial face image is input, the input image undergoes a convolution operation through a 9\texttimes9 convolutional layer, this convolutional layer extracts the basic features such as the edges and texture of the input face, and then the feature map undergoes two 3\texttimes3 convolutional layers, each of which reduces the size of the feature map while increasing the depth of the features. After passing through two convolutional layers, the feature map is normalized by three spectral normalization layers. Finally, the feature map passes through six residual blocks. Higher-level information features such as contours, details, and structures of the face are learned and the encoder structure is shown in  Fig~\ref{fig:4}. The preservation of these fundamental structures is essential for maintaining semantic consistency during generation \cite{priftiSmartContractVulnerability2024, suDiSCoDecompilingEVM2025, wei2025AdvancedSmartContract, zhuang2021smart}.

\begin{figure}[!htbp]
    \centering
    \includegraphics[width=0.8\linewidth]{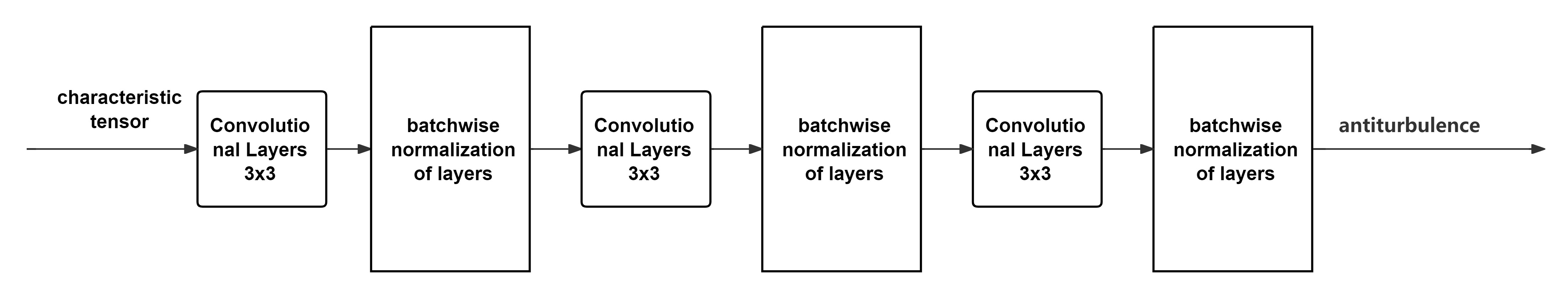}
    \caption{Decoder Structure}
    \label{fig:5}
\end{figure}

The decoder part adopts the structure of the transpose convolutional layer and batch normalization layer, which is used to decode the feature mapping extracted by the encoder part into the final image output. It consists of two 3\texttimes3 transposed convolutional layers, one 3\texttimes3 ordinary convolutional layer, and three batch normalization layers, and such a structural design ensures that the generated images have better quality and realism. The network structure is shown in Fig~\ref{fig:5}.

After the encoder converts the input face images into low-dimensional feature representations, the decoder's task is to reconstruct the antagonistic perturbations from these feature representations so that the perturbations can be added to the original image to generate antagonistic samples. The decoder employs a transposed convolutional layer to incrementally increase the feature map and eventually generate an antagonistic perturbation of the same size as the original image. The procedure is as follows:

(1) 3\texttimes3 transposed convolutional layer: this layer doubles the size of the feature map so that a higher resolution image can be recovered from the low dimensional features obtained from the encoding.

(2)  3\texttimes3 transposed convolutional layer: this layer again doubles the size of the feature map, which is closer to the size of the original image.

(3)  3\texttimes3 convolutional layer: the feature map is optimized through convolutional operations to extract and adjust features, making the final antiperturbation more suitable for combining with the original image.

(4) Batch normalization layer: A batch normalization layer is used in the decoder for normalization.

Through the above decoding process, the decoder gradually generates the antiperturbation with the same size as the original image, thus realizing the generation of perturbation on the input face image.
\subsection{Significant region extractor}
\

The structure of the salient region extractor also three convolutional layers and three batch normalization layers, the difference is that the dimension of the output of the last one convolutional layer is changed because the output of the salient mapper is usually of the same dimensions as the original image, while the output of the decoder in the generator is of the same dimensions as the input image. The salient region extractor receives the output of the encoder in the generator as input. The main process steps are:

(1) Input one feature map containing the feature information extracted after encoding the input image. The size of the feature map is increased by a factor of one through one 3\texttimes3 transposed convolutional layer.

(2) The size of the feature map is increased by  using  3\texttimes3 transposed convolutional layer again.

(3) one regular 3\texttimes3 convolutional layer is used. It keeps the size of the feature map the same, but reduces the dimensionality by decreasing the number of channels of the feature map to one, while using the BatchNorm layer for normalization.

\begin{figure}[!htbp]
    \centering
    \includegraphics[width=0.8\linewidth]{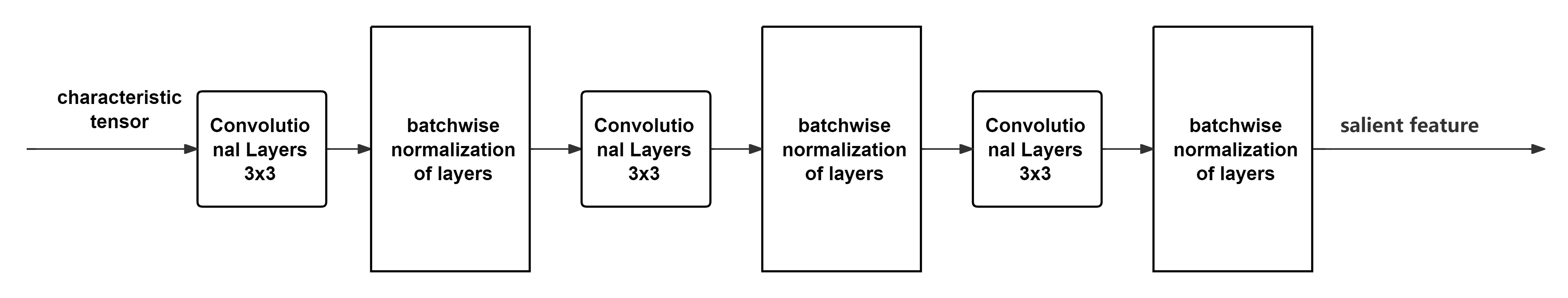}
    \caption{Structure of the salient region extractor}
    \label{fig:6}
    \vspace{-0.5em}
\end{figure}

(4) A significant map with values between zero and one is obtained through processing. In this saliency map, larger values represent higher importance in the input image, 
while smaller values represent lower importance. This saliency map indicates the saliency of different regions in the input image and provides useful information for further image analysis or processing. The structure of the salient region raiser is shown in Fig~\ref{fig:6}.
\subsection{Discriminator}
\

The discriminator uses the FaceNet model based on the Inception Resnet V1 architecture, which combines the advantages of the Inception module and the Residual Connection. The Inception module processes the input data in parallel with multiple convolutional kernels of different sizes to capture features at different scales, employing a multi-scale analysis approach similar to the opcode vectorization strategy used in smart contract security analysis. The Residual Connection, on the other hand, helps to solve the problem of gradient vanishing and representation bottleneck in deep networks, allowing the network to stack deeper without performance degradation \cite{9}. When this structure is applied to face recognition, FaceNet is able to extract richer and more robust face features. These features are more robust to changes in illumination, expression, and posture, thus improving the accuracy of face recognition.

The module works as follows:

(1) Input the confrontation sample synthesized by the synthesizer.

(2) The sample image is passed through a series of convolution, pooling, and residual modules for feature extraction and transformation of the image. The model outputs embedded features.

(3) Compare the feature vectors of the target face with the template face to determine the identity of the face object.

The FaceNet model works by converting face images into feature vectors and mapping these feature vectors into a multidimensional space. In this multidimensional space, the feature vectors corresponding to the face images of the same person are close together, while the feature vectors corresponding to the face images of different people are farther apart. By calculating the distance between these feature vectors in Euclidean space, the similarity between faces can be evaluated, with stability verification based on pattern consistency detection. If the distance between the feature vectors of two face images is less than a certain threshold, they are considered to be from the same person; otherwise, they are considered to be from different people. This method can effectively distinguish faces and achieve face recognition. The Euclidean spatial distance formula is shown below, where d represents the distance, and $x_{1m}$,$x_{2m}$ represent the feature vectors of the two images.

 \[ d = \sqrt{ \sum_{m=1}^n (x_{1m} - x_{2m})^2 } \]

\section{Experiment}\label{4}
\subsection{Experimental platform}
\ \ The experiments in this paper are realized by renting the GPU cloud-sharing platform machine of Moment Pool Cloud, and the specific experimental environment is as follows. Processor: Intel Xeon Platinum 8260C Graphics. card: NVIDIA GeForce RTX 3090. Memory: 86G . Video Memory: 24G. Operating. system: Ubantu 20.04. Software platform: Anaconda-Python 3.10. Deep Learning Framework: PyTorch 2.1.1. Main dependent libraries: CUDA 11.8 cuDNN 8

\subsection{Data presentation}
\ \ The AR Face Database (also known as AR Face Database) contains image data of 50 males and 50 females, 26 images per person, about more than 5,000 images. The face part of these images has different expression variations, such as smiling or not smiling, eyes open or not open, and wearing glasses or not. It is used as the initial training set of the model in this experiment.

The LFW (Labeled Faces in the Wild) dataset contains a set of face images from the Internet with different poses, lighting conditions, expressions, and ages. There are 13,233 face images in total. It is used for validation experiments in this experiment.
Yale Face Database: contains 165 face images from 15 different people, each containing 11 different expressions and lighting conditions. It is used for comparison experiments in this experiment.

CelebA (Celebrities Attributes) is a large-scale face attribute dataset containing about 200,000 celebrity images from the Internet. It is used for comparison experiments in this experiment.

CALFW (Caltech Faces Dataset) is a dataset for face verification tasks. It consists of about 3,000 face images from about 500 identities. It is used for comparison experiments in this experiment.
\vspace{-0.5em}
\subsection{Data processing}
\

\begin{figure}[H]
    \centering
    \includegraphics[width=0.8\linewidth]{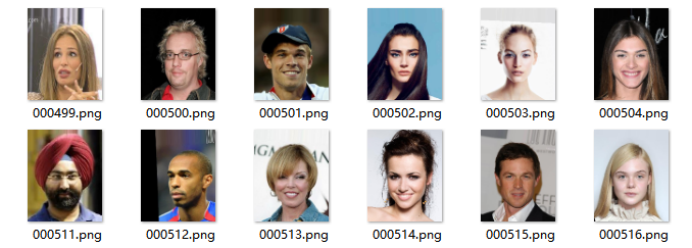}
    \caption{Sample dataset after processing}
    \label{fig:7}
\end{figure}

\begin{figure}[!htbp]
    \centering
    \includegraphics[width=0.8\linewidth]{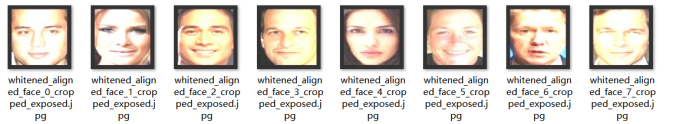}
    \caption{Whitening Processing Data}
    \label{fig:8}
\end{figure}

This paper's uses AR face library as the training set. Each image is aligned and cropped to 128\texttimes128 using the face detector MTCNN.\ Fig~\ref{fig:7} shows some of the data after MTCNN alignment and cropping. For other datasets used to perform validation and comparison experiments again each image is aligned and cropped to 128\texttimes128 using face detector MTCNN.
The AR face library dataset for GAN model training and the sample images generated by the generator are whitened before the training data enters the discriminator. The data is linearly transformed to have zero mean and unit variance. This preprocessing method can help improve the comparability of the data and the performance of the model. Fig~\ref{fig:8} illustrates the results of the image data whitening process.

\subsection{Model Training}
\subsubsection{Experimental parameters}\label{4.4.1}
\

The size of the input image in this experiment is 121\texttimes121 with pixel values between 0 and 255. The Learning Rate is set to 0.0002 and the Batch Size is set to 16. The Batch Size is set to 16 and the Epochs are set to 100.

The following loss functions are used in the generator and salient region extraction modules respectively:

(1) Mean Squared Error (MSE) loss function: used to measure the difference between the generated image and the target image. The difference between the generated image and the target image is used as part of the target attack loss, which is used to guide the generator to produce adversarial samples that are closer to the target image.

(2) Frobenius Loss: Used to constrain the magnitude of the generated adversarial perturbation to avoid the perturbation being too large. the Frobenius Loss is multiplied by a small coefficient added to the total loss as a regularization term to help control the magnitude of the perturbation.

For the judgmental FaceNet model a ternary loss function is used. An introduction to this loss function has been given in section \ref{2}.

\vspace{-11.719em}
\subsubsection{Experimental method and procedure}\label{4.4.2}
\

The experimental steps in this paper are:

(1) The initial image is fed into the synthesizer, and the high-level features in the initial image are decoded by the encoder in the synthesizer, and decoded by the decoder, and the antiperturbation factors are finally synthesized.

(2) Secondly, the significant region extraction module decodes the perturbation input from the decoder in the synthesizer and generates a feature map reflecting the importance of each part of the initial image, and combines the antagonistic perturbation factor with the feature map and combines it with the initial image to obtain the antagonistic sample.

(3) The antagonistic sample is fed into the discriminator for recognition judgment to improve the recognition ability of the discriminator. At the same time, the discriminator feeds the recognition results back to the generator part, guides the generator to generate the sample image through the mean square error, and at the same time, controls the size of the antagonistic perturbation added to the significant feature region by the significant region extraction module through the Frobenius Loss. Through the dynamic game process between the generator and the discriminator, the recognition ability of the discriminator is improved.

(4) Validation The FaceNet face recognition model trained by this paper's method is compared with the FaceNet model with the same structure obtained by training on large-scale datasets VGGface2 and CASIA-Webface on the LFW dataset in the comparison experiments, respectively, to judge the effectiveness and feasibility of this paper's experimental method, as well as the proposed viewpoints of this paper, through the introduction of the accuracy rate.

\begin{figure}[H]
    \centering
    \begin{minipage}[b]{0.8\linewidth}
        \centering
        \includegraphics[width=\linewidth]{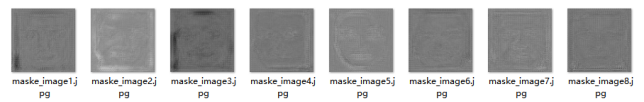}
        \caption{Counteracting disturbance factors}
        \label{fig:a}
    \end{minipage}
    
    \vspace{4pt}
    
    \begin{minipage}[b]{0.8\linewidth}
        \centering
        \includegraphics[width=\linewidth]{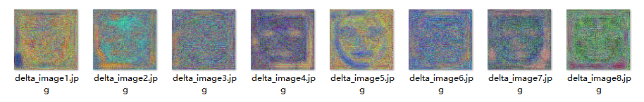}
        \caption{Salient feature extraction}
        \label{fig:b}
    \end{minipage}
    
     \vspace{4pt}
    
    \begin{minipage}[b]{0.8\linewidth}
        \centering
        \includegraphics[width=\linewidth]{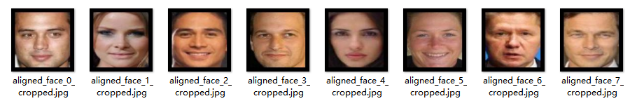}
        \caption{Salient feature extraction}
        \label{fig:c}
    \end{minipage}
\end{figure}

Fig \ref{fig:a} —\ref{fig:c} demonstrates the process of synthesizing adversarial samples by the synthesizer. Where Fig~\ref{fig:a} shows the adversarial perturbation, Fig~\ref{fig:b} shows the salient feature extraction, and Fig~\ref{fig:c} shows the generated adversarial sample. The salient map reflects the importance of each part of the face image, the higher the importance of the region in the face image corresponds to a larger value of the salient map, and the lower the importance of the region corresponds to a smaller value of the salient map. From the figure, it can be seen that the saliency map generated by this method can well determine different perturbation regions according to different inputs, and generate different weights for the antiperturbation according to the relative importance of each part of the input image, thus improving the image quality of the antiperturbation samples. With the input of about 5000 face images from the initial AR face library dataset, after the random feature perturbations are added by the GAN model generator and the salient region extraction module in this paper, finally about 500000 (0.5M) sample images are generated for the training of the discriminator.
\
\subsection{Model Implementation}  % 提升为subsection更符合文档结构
\

The implementation is based on Python with the PyTorch framework. The code structure consists of four core modules:

\begin{table}[H]
\centering
\caption{Model Code Composition}
\label{tab:code_structure}
\begin{tabular}{@{}p{0.25\textwidth}p{0.65\textwidth}@{}}
\toprule
\textbf{Code File} & \textbf{Function} \\ 
\midrule
\texttt{dataloader} & Define the methodology for all data processing in this paper \\
\texttt{FaceNet} & Constructing the discriminator for the GAN model in this paper \\
\texttt{generator} & Constructing the generators for the GAN models in this paper, with interaction-aware security checks\\
\texttt{train\_GAN} & Define generator and discriminator calls, and GAN model training methods \\ 
\bottomrule
\end{tabular}
\end{table}

Where in generator, the salient feature extraction module mentioned in this paper is implemented by defining a class named SAE, in which two decoders are included: a perturbation decoder and a mask decoder. The perturbation decoder is used to learn the transformations that perturb the input image
\begin{verbatim}
    self.perturb_decoder_1 = nn.Sequential
\end{verbatim}
the mask decoder is used to learn the generation of a mask for the reconstruction of the image 
\begin{verbatim}
    self.mask_decoder_3 = nn.Sequential
\end{verbatim}

The train-GAN file implements the training process with security monitoring, including:
(1)  Loading the data with integrity checks
(2)  Model parameter validation
(3)  Training process recording with anomaly detection
\vspace{1em}

\subsection{Experimental results and analysis}

\subsubsection{GAN model training results}
\

\begin{figure}[H]
    \centering
    \includegraphics[width=0.8\linewidth]{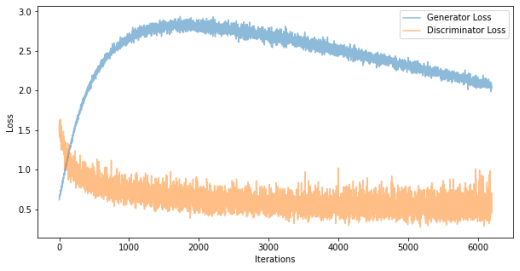}
    \caption{Loss function change during model training process}
    \label{fig:12}
    \vspace{-1.4em}
\end{figure}

Fig~\ref{fig:12} shows the change process of the generator loss function and discriminator loss function during the training of the GAN model in this paper.

The generator loss function first rises and then falls, at the beginning of training, the generator may generate some low-quality samples, which are more different from the real samples, resulting in the discriminator being more likely to recognize them as false samples, so the generator loss function may rise. As training proceeds, the generator gradually learns a better generation strategy that produces generated samples that are more realistic and closer to the real samples. This makes it more challenging for the discriminator to distinguish between the generated samples and the real samples, as the differences between them gradually become blurred. As a result, the loss function of the generator decreases gradually.

The discriminator loss function decreases as a 1/x function. In the initial stage of training, the discriminator has a more limited ability to distinguish between real and generated samples, and therefore the loss function is higher. As training proceeds, the discriminator gradually learns more accurate feature representations, which improves its ability to distinguish between real samples and generated samples, so the loss function gradually decreases. This process can be understood as the discriminator continuously improving its ability to better distinguish between different types of samples. When the performance of the discriminator is good enough, its loss function may stabilize and no longer change significantly.
\

\subsubsection{Experimental results}
\

To verify the real face recognition effect of the discriminator in section \ref{4.4.1}. The LFW dataset is utilized for verification. Choose the face recognition accuracy on the LFW dataset as the accuracy of face recognition as the validation index.

\begin{figure}[H]
    \centering
    \includegraphics[width=0.8\linewidth]{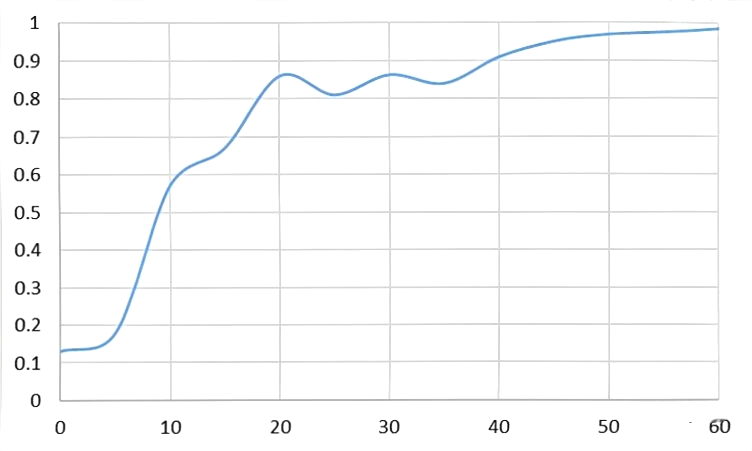}
    \caption{Model Face Recognition Accuracy}
    \label{13}
\end{figure}

The validation is performed using the LFW dataset, and the number of model iterations is set to 100. After validation, the recognition accuracy of the FaceNet face model trained using the method in this paper is shown in Fig~\ref{13}. It can be seen that the recognition accuracy is close to 100% 
when the number of iterations is 60. This result confirms the feasibility of this paper to utilize the GAN model features for small sample data enhancement and to train the face recognition model using the dynamic game features of the GAN model.
\subsubsection{Comparison Test}
\

To further validate the performance of the face recognition model trained under small sample data enhancement using GAN model features in this paper. The same structure FaceNet model trained on large datasets VGGFace, CASIA-Webface, and the FaceNet model trained by this paper's method are put on the CelebA dataset and CALFW dataset, YaleFace Database dataset for verification. In addition to the face recognition accuracy as the comparison benchmark, the number of faces required to recognize the same person is also used as the reference index for this comparison experiment. The CelebA dataset and the CALFW dataset are used as the datasets for face recognition accuracy, and the Yale Face Database dataset is used as the dataset for the number of images required to recognize the same person. The results of the comparison experiment are shown in Tab~\ref{tab:accuracy} and Tab~\ref{tab:image_requirements}.

Meanwhile, Tab~\ref{tab:training_data} shows the FaceNet model trained based on the method of this paper, FaceNet model trained based on the VGGFace dataset, and FaceNet model pre-trained based on CASIA-Webface. the number of dataset images used for the training of the three models.

% === 表1：精度比较 ===

\begin{table}[h]
\centering
\caption{Face Recognition Accuracy Comparison}
\label{tab:accuracy}
\begin{tabular}{@{} >{\raggedright}p{0.35\linewidth} C{0.3\linewidth} C{0.25\linewidth} @{}}
\toprule
\textbf{Model} & \textbf{Test Dataset} & \textbf{Accuracy (\%)} \\ 
\midrule
Our Method (FaceNet) & CelebA & 97.82 \\ 
                     & CALFW  & 98.93 \\ 
\addlinespace[2pt]
VGGFace-based & CelebA & 99.92 \\
              & CALFW  & 99.94 \\
\addlinespace[2pt] 
CASIA-Webface & CelebA & 97.52 \\
              & CALFW  & 98.71 \\
\bottomrule
\end{tabular}
\end{table}

% === 表2：图像需求 ===
\begin{table}[h]
\centering
\caption{Minimum Image Requirements for Recognition}
\label{tab:image_requirements}
\begin{tabular}{@{} L{0.5\linewidth} C{0.3\linewidth} C{0.15\linewidth} @{}}
\toprule
\textbf{Model} & \textbf{Test Dataset} & \textbf{Images Required} \\ 
\midrule
FaceNet (Our Method) & YaleFace & 13 \\ 
FaceNet (VGGFace) & YaleFace & 10 \\
FaceNet (CASIA) & YaleFace & 32 \\
\bottomrule
\end{tabular}
\end{table}
% === 表3：训练数据 ===
\begin{table*}[h]  % 使用table*跨双栏
\centering
\caption{Training Dataset Specifications}
\label{tab:training_data}
\begin{tabular}{@{} L{0.45\textwidth} L{0.35\textwidth} C{0.15\textwidth} @{}}
\toprule
\textbf{Model} & \textbf{Training Dataset} & \textbf{Image Count} \\ 
\midrule
FaceNet (Our Method) 
& AR Face Library + Generated Samples 
& 0.5M+ \\

\addlinespace[3pt]
FaceNet (VGGFace) 
& VGGFace Dataset 
& 2.6M \\

\addlinespace[3pt]
FaceNet (CASIA) 
& CASIA-Webface Dataset 
& 0.6M \\
\bottomrule
\end{tabular}
\end{table*}

\subsubsection{Analysis of Experimental Results}
\

The validation experiments in section \ref{4.4.2} show that this paper utilizes the knowledge of residual networks to construct the GAN model with good results, which effectively avoids the problems of gradient disappearance and explosion encountered by the ordinary GAN model in the training process. We successfully utilize the dynamic game characteristics of the GAN model, train FaceNet as the discriminator of the GAN model in this paper, and finally test it in the LFW dataset to achieve good recognition results.

Meanwhile, the comparison experiments in section \ref{4.4.2} show that this paper also achieves good results using the GAN model for small-sample data enhancement, and in the final experimental comparison of the face recognition rate and the images needed to recognize the same person with the same structural model trained based on a large dataset, the FaceNet face recognition trained using the method adopted in this paper also has a good performance.

\section{Future work}
\

The next step will be to explore the possibilities of this paper's method for cross-age-based face recognition. This is of great significance for the identification of missing children, matching the childhood photos of the missing children with their adult photos, even if the children's appearance has changed greatly after they have grown up, they can still be found by technical means. At the same time, dynamic face recognition is also the next direction of this paper's efforts, real-life face recognition is usually carried out in dynamic environments, and at the same time, the use of AI face-switching technology to carries out fraudulent video cheating behavior is also an application of dynamic face recognition.

By utilizing the existing methods, we explore the possibility of realizing dynamic face recognition as well as contend with the prevention of AI face-swapping technology. At the same time, we design relevant systems to deploy the trained face recognition models to relevant hardware devices and apply them to real life. In addition to using the FaceNet model as a GAN model discriminator, we will explore the effect of other face recognition models as GAN model discriminators.

\section{Conclusion}\label{5}
\

Through several months of related literature reading and experiments, we realized the construction of the GAN model using the residual network and solved the problems of gradient disappearance and explosion encountered in the training process of the GAN model. At the same time contend with the face recognition model under complex conditions, the training samples are difficult to obtain, and the lack of sufficient samples leads to poor model robustness and generalization ability. In this paper, the possibility of small sample data enhancement using the GAN model. The significant feature extraction module is added to the GAN model realized by the residual network to extract the significant features of the face and randomly add perturbations to it. Small sample data enhancement is successfully performed to expand the face recognition model training set. Using the dynamic game characteristics of the GAN model, the FaceNet model based on the Inception Resnet V1 structure is used as the discriminator of the GAN, which further improves the recognition ability of the face recognition model. Finally, the feasibility of the work done in this paper is verified through validation experiments and comparison experiments \cite{aguilarSmartContractFamilies2024, arceriSoundConstructionEVM2024, ayubSoundAnalysisMigration2024, chenDemystifyingInvariantEffectiveness2024, grossmanPracticalVerificationSmart2024}.

The work in this paper also has deficiencies and areas for improvement. For the data enhancement in a single way, it is only realized by randomly adding antagonistic perturbations in the region of significant features. For other more complex environments, such as angle change slight occlusion, and other complex conditions, this paper has not designed relevant experiments and solutions. Meanwhile, the recognition of dynamic faces is also a place not yet considered in this work, real-life face recognition is usually carried out in dynamic environments.

\section*{Acknowledgments} 
The author would like to express sincere gratitude to Prof. Tang Wan for her invaluable
guidance and support during the foundational stages of this research.
\bibliographystyle{unsrt}
\bibliography{reference}

@inproceedings{9724923,
  author    = {Wang, Haoyu and Guo, Lulu},
  booktitle = {2021 3rd International Conference on Artificial Intelligence and Advanced Manufacture (AIAM)},
  title     = {Research on Face Recognition Based on Deep Learning},
  year      = {2021},
  volume    = {},
  number    = {},
  pages     = {540-546},
  doi       = {10.1109/AIAM54119.2021.00113}
}

@article{adjabi2020past,
  title     = {Past, present, and future of face recognition: A review},
  author    = {Adjabi, Insaf and Ouahabi, Abdeldjalil and Benzaoui, Amir and Taleb-Ahmed, Abdelmalik},
  journal   = {Electronics},
  volume    = {9},
  number    = {8},
  pages     = {1188},
  year      = {2020},
  publisher = {MDPI}
}

@article{mehta2022three,
  title     = {Three-dimensional DenseNet self-attention neural network for automatic detection of student’s engagement},
  author    = {Mehta, Naval Kishore and Prasad, Shyam Sunder and Saurav, Sumeet and Saini, Ravi and Singh, Sanjay},
  journal   = {Applied Intelligence},
  volume    = {52},
  number    = {12},
  pages     = {13803--13823},
  year      = {2022},
  publisher = {Springer}
}

@inproceedings{schroff2015facenet,
  title     = {Facenet: A unified embedding for face recognition and clustering},
  author    = {Schroff, Florian and Kalenichenko, Dmitry and Philbin, James},
  booktitle = {Proceedings of the IEEE conference on computer vision and pattern recognition},
  pages     = {815--823},
  year      = {2015}
}

@article{liu2023deep,
  title     = {Deep learning based single sample face recognition: a survey},
  author    = {Liu, Fan and Chen, Delong and Wang, Fei and Li, Zewen and Xu, Feng},
  journal   = {Artificial Intelligence Review},
  volume    = {56},
  number    = {3},
  pages     = {2723--2748},
  year      = {2023},
  publisher = {Springer}
}

@article{alqahtani2021applications,
  title     = {Applications of generative adversarial networks (gans): An updated review},
  author    = {Alqahtani, Hamed and Kavakli-Thorne, Manolya and Kumar, Gulshan},
  journal   = {Archives of Computational Methods in Engineering},
  volume    = {28},
  pages     = {525--552},
  year      = {2021},
  publisher = {Springer}
}

@article{kammoun2022generative,
  title     = {Generative adversarial networks for face generation: A survey},
  author    = {Kammoun, Amina and Slama, Rim and Tabia, Hedi and Ouni, Tarek and Abid, Mohmed},
  journal   = {ACM Computing Surveys},
  volume    = {55},
  number    = {5},
  pages     = {1--37},
  year      = {2022},
  publisher = {ACM New York, NY}
}

@article{9,
  author  = {  Guoxiang Tong and     Fangning Hu and Hongjun Liu},
  title   = {DAGAN: A GAN Network for Image Denoising of Medical Images Using Deep Learning of Residual Attention Structures},
  journal = {International Journal of Pattern Recognition and Artificial Intelligence},
  volume  = {38},
  number  = {02},
  year    = {2024},
  issn    = {0218-0014}
}

@article{li2020review,
  title     = {A review of face recognition technology},
  author    = {Li, Lixiang and Mu, Xiaohui and Li, Siying and Peng, Haipeng},
  journal   = {IEEE access},
  volume    = {8},
  pages     = {139110--139120},
  year      = {2020},
  publisher = {IEEE}
}

@article{wu2021mtcnn,
  title     = {MTCNN and FACENET based access control system for face detection and recognition},
  author    = {Wu, Chunming and Zhang, Ying},
  journal   = {Automatic Control and Computer Sciences},
  volume    = {55},
  pages     = {102--112},
  year      = {2021},
  publisher = {Springer}
}

@article{cui2019image,
  title     = {Image data augmentation for SAR sensor via generative adversarial nets},
  author    = {Cui, Zongyong and Zhang, Mingrui and Cao, Zongjie and Cao, Changjie},
  journal   = {IEEE Access},
  volume    = {7},
  pages     = {42255--42268},
  year      = {2019},
  publisher = {IEEE}
}

@article{yan2021enhanced,
  title     = {Enhanced network optimized generative adversarial network for image enhancement},
  author    = {Yan, Lingyu and Fu, Jiarun and Wang, Chunzhi and Ye, Zhiwei and Chen, Hongwei and Ling, Hefei},
  journal   = {Multimedia Tools and Applications},
  volume    = {80},
  pages     = {14363--14381},
  year      = {2021},
  publisher = {Springer}
}

@article{wang2021generative,
  title     = {Generative adversarial networks in computer vision: A survey and taxonomy},
  author    = {Wang, Zhengwei and She, Qi and Ward, Tomas E},
  journal   = {ACM Computing Surveys (CSUR)},
  volume    = {54},
  number    = {2},
  pages     = {1--38},
  year      = {2021},
  publisher = {ACM New York, NY, USA}
}

@article{hwang2023adversarial,
  title     = {Adversarial patch attacks on deep-learning-based face recognition systems using generative adversarial networks},
  author    = {Hwang, Ren-Hung and Lin, Jia-You and Hsieh, Sun-Ying and Lin, Hsuan-Yu and Lin, Chia-Liang},
  journal   = {Sensors},
  volume    = {23},
  number    = {2},
  pages     = {853},
  year      = {2023},
  publisher = {MDPI}
}

@article{banerjee2018lr,
  title     = {LR-GAN for degraded face recognition},
  author    = {Banerjee, Samik and Das, Sukhendu},
  journal   = {Pattern Recognition Letters},
  volume    = {116},
  pages     = {246--253},
  year      = {2018},
  publisher = {Elsevier}
}

@inproceedings{wu2025atomicity,
  title   = {On the Atomicity and Efficiency of Blockchain Payment Channels},
  author  = {Wu, Di and Ren, Shoupeng and Bai, Yuman and He, Lipeng and Liu, Jian and Wen, Wu and Ren, Kui and Chen, Chun},
  journal = {Proceedings of the 34th USENIX Security Symposium (USENIX Security 25)},
  pages   = {4053--4072},
  year    = {2025}
}

@inproceedings{xiao2025parallelizing,
  title     = {Parallelizing Universal Atomic Swaps for $\{$Multi-Chain$\}$ Cryptocurrency Exchanges},
  author    = {Xiao, Danlei and Zhang, Chuan and Deng, Haotian and Liang, Jinwen and Wang, Licheng and Zhu, Liehuang},
  booktitle = {Proceedings of the 34th USENIX Security Symposium (USENIX Security 25)},
  pages     = {4073--4092},
  year      = {2025}
}

@inproceedings{aguilarSmartContractFamilies2024,
  title     = {Smart {{Contract Families}} in {{Solidity}}},
  booktitle = {Proceedings of the 34th {{International Conference}} on {{Collaborative Advances}} in {{Software}} and {{COmputiNg}} (CASCON)},
  author    = {Aguilar, Julio and Bak, Kacper and Boyle, Michael and Callens, Valerian and Gorzny, Jan},
  year      = 2024,
  pages     = {1--5}
}

@inproceedings{arceriSoundConstructionEVM2024,
  title     = {Towards a {{Sound Construction}} of {{EVM Bytecode Control-Flow Graphs}}},
  booktitle = {Proceedings of the 26th {{ACM International Workshop}} on {{Formal Techniques}} for {{Java-like Programs}} (FTfJP)},
  author    = {Arceri, Vincenzo and Merenda, Saverio Mattia and Dolcetti, Greta and Negrini, Luca and Olivieri, Luca and Zaffanella, Enea},
  year      = 2024,
  pages     = {11--16}
}

@article{ayubSoundAnalysisMigration2024,
  title   = {Sound Analysis and Migration of Data from {{Ethereum}} Smart Contracts},
  author  = {Ayub, Maha and Khan, Muhammad Waiz and Janjua, Muhammmad Umar},
  year    = 2024,
  journal = {Automated Software Engineering},
  volume  = {31},
  number  = {1},
  pages   = {21}
}

@inproceedings{caiEnablingCompleteAtomicity2024,
  title     = {Enabling {{Complete Atomicity}} for {{Cross-Chain Applications Through Layered State Commitments}}},
  booktitle = {Proceedings of the 43rd {{International Symposium}} on {{Reliable Distributed Systems}} ({{SRDS}})},
  author    = {Cai, Yuandi and Cheng, Ru and Zhou, Yifan and Zhang, Shijie and Xiao, Jiang and Jin, Hai},
  year      = 2024,
  pages     = {248--259}
}

@article{chenDemystifyingInvariantEffectiveness2024,
  title   = {Demystifying {{Invariant Effectiveness}} for {{Securing Smart Contracts}}},
  author  = {Chen, Zhiyang and Liu, Ye and Beillahi, Sidi Mohamed and Li, Yi and Long, Fan},
  year    = 2024,
  journal = {Reproduction package of the paper "Demystifying Invariant Effectiveness for Securing Smart Contracts"},
  volume  = {1},
  number  = {FSE},
  pages   = {79:1772--79:1795}
}

@article{wang2024ContractsentryStaticAnalysis,
  title   = {Contractsentry: A Static Analysis Tool for Smart Contract Vulnerability Detection},
  author  = {Wang, Shiji and Zhao, Xiangfu},
  year    = 2024,
  journal = {Automated Software Engineering},
  volume  = {32},
  number  = {1},
  pages   = {1}
}

@article{grossmanPracticalVerificationSmart2024,
  title   = {Practical {{Verification}} of {{Smart Contracts}} Using {{Memory Splitting}}},
  author  = {Grossman, Shelly and Toman, John and Bakst, Alexander and Arora, Sameer and Sagiv, Mooly and Nandi, Chandrakana},
  year    = 2024,
  journal = {Artifact for our paper titled "Practical Verification Of Smart Contracts using Memory Splitting"},
  volume  = {8},
  number  = {OOPSLA2},
  pages   = {356:2402--356:2433}
}

@article{hanOSwapPreservingAtomicity2026,
  title   = {{{OSwap}}: {{Preserving}} the {{Atomicity}} and {{Indistinguishability}} of \textbackslash bm N\_\textbackslash bm 1\textbackslash bm \textbackslash sim \textbackslash bm N\_\textbackslash bm 2n1{$\sim$}n2 {{Swap Without Synchronous Blockchain Communication}}},
  author  = {Han, Panpan and Yan, Zheng and Yang, Laurence T. and Bertino, Elisa},
  year    = 2026,
  journal = {IEEE Transactions on Dependable and Secure Computing},
  volume  = {23},
  number  = {1},
  pages   = {477--490}
}

@article{jiaoSurveyEthereumSmart2024,
  title   = {A {{Survey}} of {{Ethereum Smart Contract Security}}: {{Attacks}} and {{Detection}}},
  author  = {Jiao, Tengyun and Xu, Zhiyu and Qi, Minfeng and Wen, Sheng and Xiang, Yang and Nan, Gary},
  year    = 2024,
  journal = {Distributed Ledger Technologies: Research and Practice},
  volume  = {3},
  number  = {3},
  pages   = {23:1--23:28}
}

@inproceedings{kumarVulnerabilitiesSmartContracts2024,
  title     = {``{{Vulnerabilities}} in {{Smart Contracts}}: {{A Detailed Survey}} of {{Detection}} and {{Mitigation Methodologies}}''},
  booktitle = {Proceedings of the 2024 {{International Conference}} on {{Emerging Technologies}} in {{Computer Science}} for {{Interdisciplinary Applications}} ({{ICETCS}})},
  author    = {Kumar, Nayantara K and Honnungar, Niranjan V and Sharwari Prakash, M and Lohith, J J},
  year      = 2024,
  pages     = {1--7}
}

@article{liASTRODetectingAccess2025,
  title   = {{{ASTRO}}: {{Detecting Access Control Vulnerabilities}} in {{Smart Contracts}} via {{Graph Similarity Comparison}}},
  author  = {Li, Wei and Nan, Yuhong and Ye, Mingxi and Zhang, Jingwen and Zheng, Peilin and Zheng, Zibin},
  year    = 2025,
  journal = {IEEE Transactions on Software Engineering},
  volume  = {51},
  number  = {12},
  pages   = {3267--3283}
}

@inproceedings{liDemoEnhancingSmart2024,
  title     = {Demo: {{Enhancing Smart Contract Security Comprehensively}} through {{Dynamic Symbolic Execution}}},
  booktitle = {Proceedings of the 2024 on {{ACM SIGSAC Conference}} on {{Computer}} and {{Communications Security}} (CCS)},
  author    = {Li, Zhaoxuan and Zhao, Ziming and Li, Wenhao and Zhang, Rui and Xue, Rui and Lu, Siqi and Zhang, Fan},
  year      = 2024,
  pages     = {5072--5074}
}

@inproceedings{priftiSmartContractVulnerability2024,
  title     = {Smart {{Contract Vulnerability Detection Using Deep Learning Algorithms}} on {{EVM}} Bytecode},
  booktitle = {Proceedings of the 13th {{Mediterranean Conference}} on {{Embedded Computing}} ({{MECO}})},
  author    = {Prifti, Lejdi and Cico, Betim and Karras, Dimitrios},
  year      = 2024,
  pages     = {1--7}
}

@article{suDiSCoDecompilingEVM2025,
  title   = {{{DiSCo}}: {{Towards Decompiling EVM Bytecode}} to {{Source Code}} Using {{Large Language Models}}},
  author  = {Su, Xing and Liang, Hanzhong and Wu, Hao and Niu, Ben and Xu, Fengyuan and Zhong, Sheng},
  year    = 2025,
  journal = {Proceedings of the ACM on Software Engineering},
  volume  = {2},
  number  = {FSE},
  pages   = {FSE103:2311--FSE103:2334}
}

@article{wangContractCheckCheckingEthereum2024,
  title   = {{{ContractCheck}}: {{Checking Ethereum Smart Contracts}} in {{Fine-Grained Level}}},
  author  = {Wang, Xite and Tian, Senping and Cui, Wei},
  year    = 2024,
  journal = {IEEE Transactions on Software Engineering},
  volume  = {50},
  number  = {7},
  pages   = {1789--1806}
}

@article{wangEfficientlyDetectingReentrancy2024,
  title   = {Efficiently {{Detecting Reentrancy Vulnerabilities}} in {{Complex Smart Contracts}}},
  author  = {Wang, Zexu and Chen, Jiachi and Wang, Yanlin and Zhang, Yu and Zhang, Weizhe and Zheng, Zibin},
  year    = 2024,
  journal = {Proceedings of the ACM on Software Engineering},
  volume  = {1},
  number  = {FSE},
  pages   = {8:161--8:181}
}

@article{weiSurveyQualityAssurance2024,
  title   = {Survey on {{Quality Assurance}} of {{Smart Contracts}}},
  author  = {Wei, Zhiyuan and Sun, Jing and Zhang, Zijian and Zhang, Xianhao and Yang, Xiaoxuan and Zhu, Liehuang},
  year    = 2024,
  journal = {ACM Computing Surveys},
  volume  = {57},
  number  = {2},
  pages   = {32:1--32:36}
}

@article{zhuSurveySecurityAnalysis2024,
  title   = {A {{Survey}} on {{Security Analysis Methods}} of {{Smart Contracts}}},
  author  = {Zhu, Huijuan and Yang, Lei and Wang, Liangmin and Sheng, Victor S.},
  year    = 2024,
  journal = {IEEE Transactions on Services Computing},
  volume  = {17},
  number  = {6},
  pages   = {4522--4539}
}

@article{wei2025AdvancedSmartContract,
  title   = {Advanced {{Smart Contract Vulnerability Detection}} via {{LLM-Powered Multi-Agent Systems}}},
  author  = {Wei, Zhiyuan and Sun, Jing and Sun, Yuqiang and Liu, Ye and Wu, Daoyuan and Zhang, Zijian and Zhang, Xianhao and Li, Meng and Liu, Yang and Li, Chunmiao and Wan, Mingchao and Dong, Jin and Zhu, Liehuang},
  year    = 2025,
  journal = {IEEE Transactions on Software Engineering},
  volume  = {51},
  number  = {10},
  pages   = {2830--2846}
}

@inproceedings{zhuang2021smart,
  title     = {Smart contract vulnerability detection using graph neural networks},
  author    = {Zhuang, Yuan and Liu, Zhenguang and Qian, Peng and Liu, Qi and Wang, Xiang and He, Qinming},
  booktitle = {Proceedings of the twenty-ninth international conference on international joint conferences on artificial intelligence (IJCAI)},
  pages     = {3283--3290},
  year      = {2021}
}

\end{document}